\documentclass{CVIS}

\usepackage{amsmath,bm,url}

\begin{document}

\title{Opening the Black Box of Financial AI with CLEAR-Trade: A CLass-Enhanced Attentive Response Approach for Explaining and Visualizing Deep Learning-Driven Stock Market Prediction}

\author{
\begin{tabularx}{\textwidth}{X X}
Devinder Kumar & University of Waterloo, ON, Canada \\
Graham W. Taylor & University of Guelph, CIFAR and Vector Institute, ON, Canada\\
Alexander Wong & University of Waterloo \& Waterloo AI Institute, ON, Canada \\
\end{tabularx}
}

\maketitle

\begin{abstract}
Deep learning have been shown to outperform traditional machine learning algorithms across a wide range of problem domains. However, current deep learning algorithms are essentially uninterpretable "black-boxes" without any explanations associated with their decision making processes. This is a major shortcoming that prevents widespread use of deep learning to be used in scenarios with regulatory processes such as finance.  As such, industries such as finance have to rely on traditional models like decision trees that are much more interpretable but less effective than deep learning for complex problems. In this paper, we propose CLEAR-Trade, a novel financial AI visualization framework for deep learning-driven stock market prediction that mitigates the interpretability issue of deep learning. In particular, CLEAR-Trade provides a effective way to visualize and explain decisions made by deep stock market prediction models. We show the efficacy of CLEAR-Trade in enhancing the interpretability of stock market prediction by conducting experiments based on S\&P 500 stock index prediction. The results clearly demonstrate that CLEAR-Trade can provide significant insight into the decision-making process of the deep learning-driven financial models, particularly for regulatory processes, thus improving their potential widespread adoption in finance.

\end{abstract}

\section{Introduction}
Do machine learning algorithms need to be explainable? This is an important question in today's world where machine learning algorithms, especially those based on deep learning are being used at a wide range of tasks and have shown tremendous efficacy in performing these tasks. Deep learning has touted as being very disruptive to many sectors, particularly the finance sector. However, deep learning, to large extent, have essentially been unexplainable "black boxes", with no clear explanation as to how they reach  particular decisions~\cite{yamins2016using}. This is a major hindrance to the widespread adoption of deep learning in industries like finance, where regulations are very tight. In such industries with strict regulatory processes, the AI models used are required to be transparent, interpretable, and explainable. Many experts in these sectors believe that relying on such 'black box' methods is a growing problem that is already very relevant due to regulatory processes in these sectors, and it is going to be increasingly more relevant in the future. For example, in finance, law requires companies to explain the reason behind every decision to its perspective customer~\cite{mitreview2017}. As such, current approaches for leveraging deep learning are not feasible such in these scenarios.

The limitation of deep learning in terms of transparency and interpretability have forced industries dealing with regulatory scenarios to use comparatively simple machine learning algorithms such as linear or logistic regression, decision trees, or ensemble methods such as random forests which are significantly more explainable and quite effective in simple cases. However, as the complexity of the problem increases, which is very true in finance, deep learning algorithms have been shown to outperform such traditional algorithms by a wide margin across a wide range of problem domains~\cite{lecun2015deep}. As such, strategies for explaining the decisions made by deep learning algorithms are highly desired to enable their widespread use in sectors that have strong regulatory processes.

More recently, a number of methods were proposed to mitigate this issue of interpretability and transparency in deep learning. For example, Zeiler \& Fergus~\cite{zeiler2014visualizing} proposed the formation of a parallel deconvolutional network to peer into different units of the network.  Ribeiro~\cite{ribeiro2016should} introduced a method to build trust in models that are locally accurate, i.e., it is correct near the input data sample. Selvaraju et.al.~\cite{selvaraju2017grad} proposed a method called Grad-CAM that enables users to discern "strong" networks from the weaker ones. While promising, all of the aforementioned approaches are restricted to identifying regions of interest and their influence in the decision made by the deep neural network only, thus restricting their utility for gaining a more detailed understanding of the decision process. To address this issue, Kumar et. al.~\cite{kumar2017explaining} recently proposed a \textbf{CL}ass \textbf{E}nhanced \textbf{A}ttentive \textbf{R}esponse (CLEAR) approach that not only identifies attentive regions of interest and their influence on the decision made, but more important provides the dominant classes associated with the attentive regions of interest. This additional information about the dominant classes and their influence on the decision making progress leads to a higher degree of human interpretability, which makes it very well suited for scenarios that necessitate regulatory processes such as in finance.

\begin{figure}
\begin{center}
   \includegraphics[trim = 0cm 0cm 0cm 0cm ,height = 7.cm,width=1\linewidth]{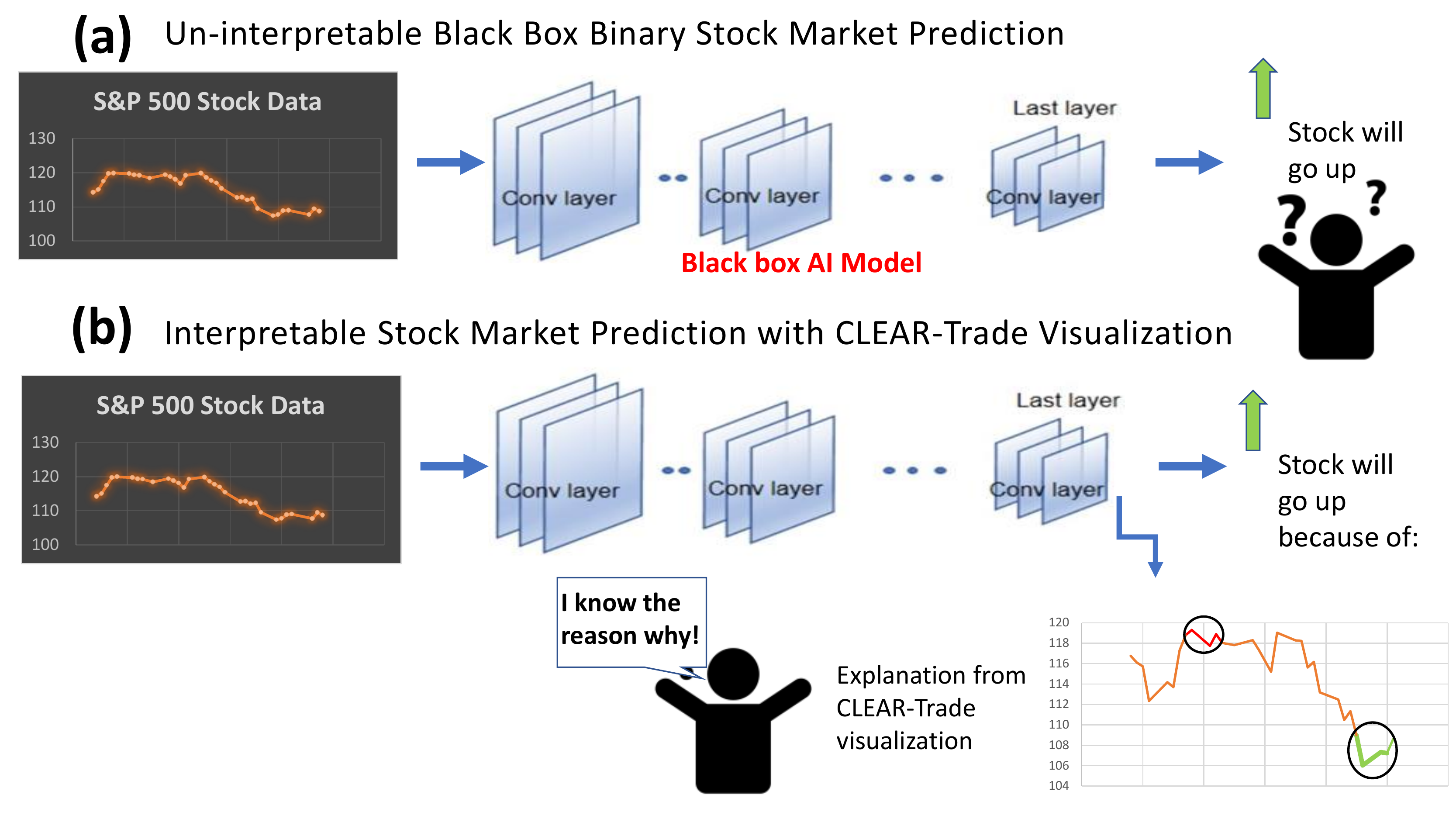}
\end{center}
   \caption{Two different scenarios for stock market prediction using deep learning-driven financial AI models: a) Stock market prediction without interpretability, b)interpretable stock market prediction via CLEAR-Trade. The proposed CLEAR-Trade visualization framework improves financial model interpretability by providing effective visual interpretations of the decision-making process. CLEAR-Trade allows for the visualization of i) the attentive time windows responsible for stock market prediction decisions based on the financial AI model (marked in red and green), ii) their level of contribution to the stock market prediction decision (in this case, stock market index rise (green) or stock market index fall (red)), as well as iii) the dominant state (rise or fall) associated with each attentive time window. This visualization enables financial analysts to better understand the rationale behind the stock market prediction decisions made based by the deep learning-driven financial AI model.}

\label{fig:motivation}
\end{figure}

Motivated by this, in this paper, we propose CLEAR-Trade, a \textbf{CL}ass \textbf{E}nhanced \textbf{A}ttentive \textbf{R}esponse approach to explaining and visualizing deep learning-driven stock market prediction.  In particular, CLEAR-Trade is designed in this paper to provide detailed explanations for the prediction decisions made by a deep learning-driven binary stock market prediction network, as shown in Fig.~\ref{fig:motivation}. Our aim is to create a powerful tool for peering into the minds of these otherwise uninterpretable 'black box' financial AI models to better visualize and understand why they are making the decisions the way they do. Doing this will have a tremendous impact on day-to-day work of financial analysts in helping them better understand these deep learning-driven financial AI models, thus potentially enabling the widespread adoption of transparent financial AI.

\begin{figure}[t]
\begin{center}
   \includegraphics[trim = 0cm 0cm 0cm 0cm ,height = 7cm,width=1\linewidth]{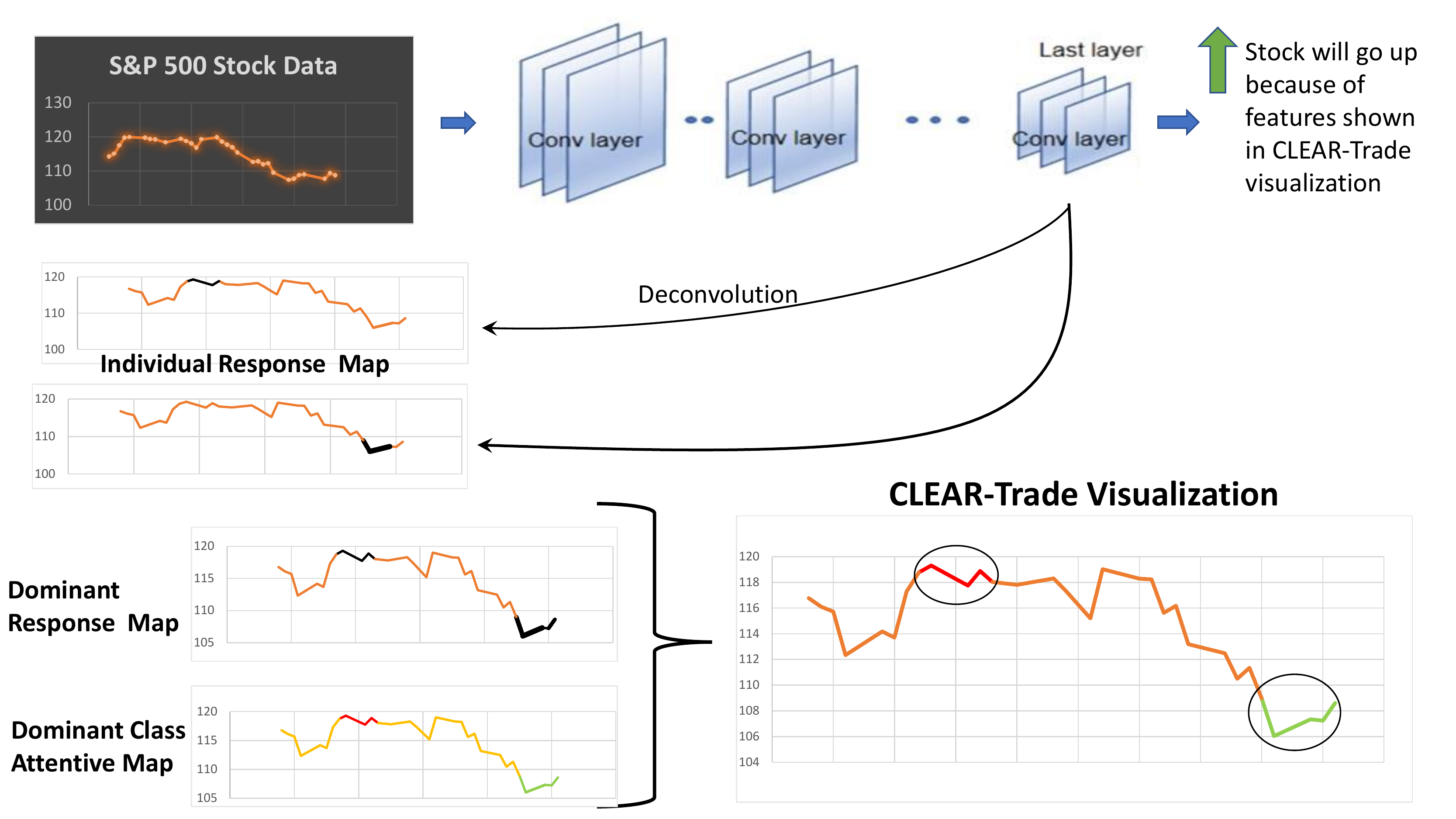}
\end{center}
   \caption{An overview of the proposed CLEAR-Trade visualization framework for explaining and visualizing deep learning-driven stock market prediction. As an illustrative case in this paper, for predicting whether a stock market index will rise or fall (two states), the index's past 30 days of trade data is fed into the deep learning-driven financial AI model and individual attentive response maps are computed for each state (stock market index rise or stock market index down) based on the last layer of the deep learning-driven financial AI model. Based on this set of attentive response maps, two different maps are computed: 1) a dominant attentive response map, which shows the level of contribution of each time point to the decision-making process, and 2) a dominant state attentive map, which shows the dominant state associated with each time point influencing the decision-making process. Finally, the dominant attentive response map and the dominant attentive state map are combined to produce the final CLEAR-Trade visualization, thus enabling the financial analyst to visualize the factors leveraged by the deep learning-driven financial AI model in predicting whether the stock market index will rise (green) or fall (red).}

\label{fig:clear}
\end{figure}

\section{Methodology}
\label{method}
With the goal of enabling transparent and interpretable deep learning-driven stock market prediction, the proposed CLEAR-Trade visualization framework presents the financial analyst with the following information pertaining to the decision-making process:
\begin{enumerate}
\item the attentive time windows responsible for the decision made by the financial AI model; 
\item the attentive levels at these attentive time windows so that their level of influence over the decision made by the financial AI model can be understood; and 
\item the dominant state (in this paper, stock market index rise or fall) associated with these attentive time windows so that we can better understand why a decision was made. 
\end{enumerate} 

The procedure for obtaining the CLEAR-Trade visualization for stock market prediction (in this case, predicting stock market indices but can also be applied to individual stocks) is shown in Fig.~\ref{fig:clear} and can be explained as follows.

First, a forward pass with a time-series input of historical trade information about a particular stock market index (in this case, an index's 30 days worth of open, close, highs, lows, and trade volumes) is performed through the deep learning-driven financial AI model and a stock market prediction decision output is obtained. To create a CLEAR-Trade visualization associated with this particular stock market prediction decision, we first compute a set of individual response maps  $\left\{R(\underline{x}\lvert s)|1 \leq s \leq K\right\}$, where $K$ is the total number of states present for stock market prediction (in this case, there are two states: stock market index rise and fall). The deep learning-driven financial AI model is set up such that it contains similar number of kernels in the last layer as the number of states. To elaborate the process, first consider the response for all the kernels at the the last layer $l$ of the financial AI model which can be calculated as:

\begin{equation}
 \hat h_{l} = \sum_{k=1}^K z_{k,l} * w_{k,l} .
\end{equation}

\noindent where $*$ denotes the convolution operation. To calculate the response of last layer in the input domain, we can extend this formulation for response of the specific kernel $s\epsilon\{1...K\}$ of the deep learning-driven financial AI model with Un-pooling layer $P'$ as:
\begin{equation}
 {R(\underline{x}\lvert s)} = H_{1}P'_{1}H_{2}P'_{2} ....H_{L-1}P'_{L-1}H_{L}^s z_{L}.
\end{equation}

\noindent where $H$ denotes the combined operation of convolutional and summation, for notation brevity. $H_{L}^s$ represents the convolution matrix operation in which the kernel weights $w_{L}$ are all zero except that at the $s$\textsuperscript{th} time point.

Given the set of individual attentive response maps, we then compute the dominant attentive state map, $\hat{S}(\underline{x})$, by finding the state that maximizes the attentive response level, $R(\underline{x}\lvert s)$, across all states:

\begin{equation}
\vspace{-0.2cm}
 \hat{S}(\underline{x}) = \operatornamewithlimits{argmax}\limits_{s} {R(\underline{x} \lvert s)} .
\end{equation}

Given the dominant attentive state map, $\hat{S}(\underline{x})$, we can now compute the dominant attentive response map, $D_{\hat{S}}(\underline{x})$, by selecting the attentive response level at a particular time point based on the identified dominant state, which can be expressed as follows:

\begin{equation}
 D_{\hat{S}}(\underline{x}) = R(\underline{x}\lvert \hat{S}) .
\end{equation}

To form the final CLEAR-Trade visualization, we map the dominant state attentive map and the dominant attentive response map in the HSV (S in HSV is indicated as S' below to avoid confusion with state $S$) color space as follows:

\begin{equation}
\begin{split}
   H & = F(\hat{S}(\underline{x})) ,\ \\
   S' & = 1 ,\ \\
   V & = D_{\hat{S}}(\underline{x}) .\
\end{split}
\end{equation}

\noindent where $F(.)$ is the color map dictionary that assigns an individual color to each dominant attentive state, $s$. Fig.~\ref{fig:clear} shows an example of the CLEAR-Trade visualization.

\begin{figure}[t]
\begin{center}
   \includegraphics[trim = 0cm 10cm 4cm 4cm ,height = 2.5cm,width=1\linewidth]{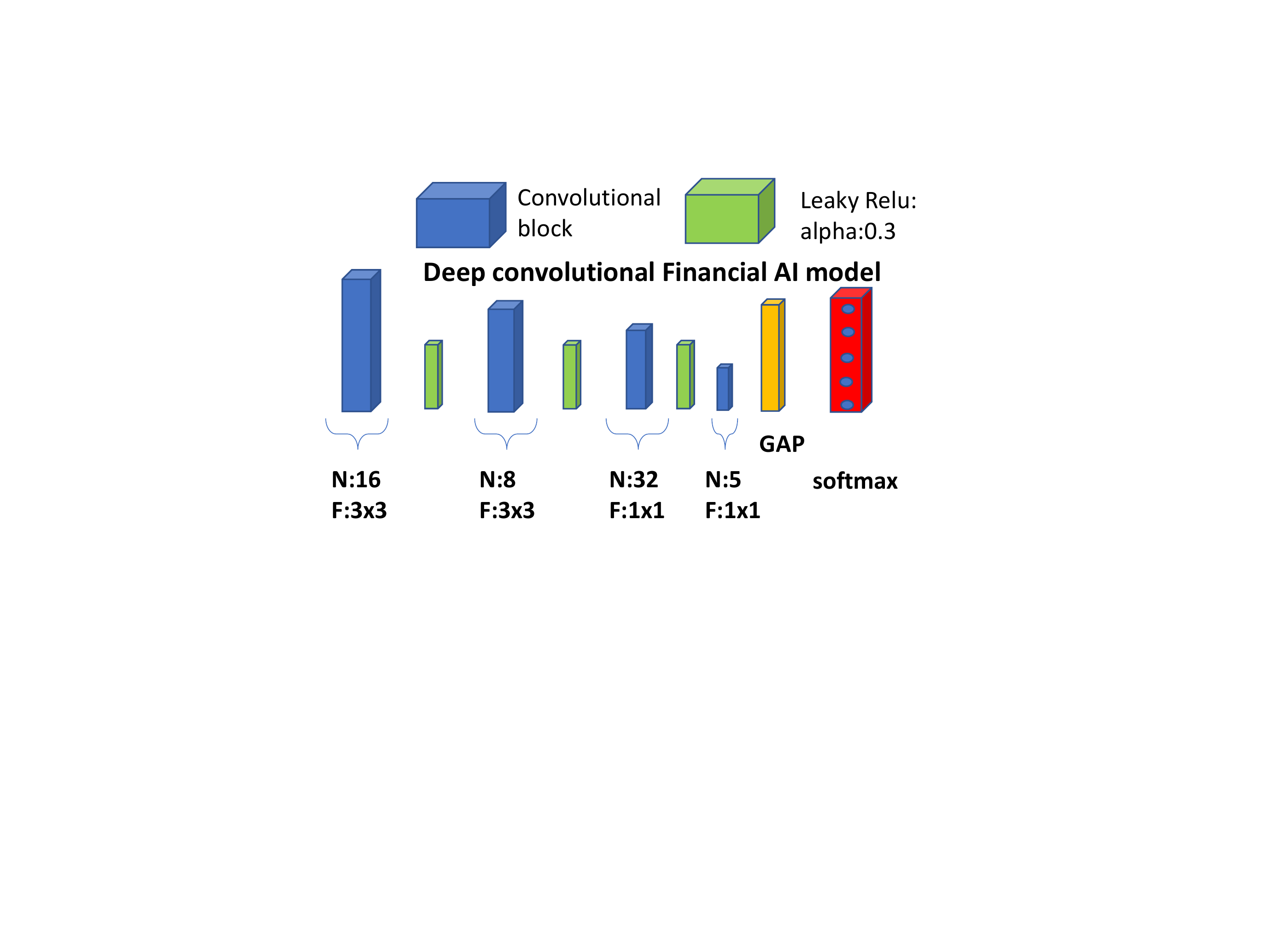}
\end{center}

   \caption{Architecture of the deep convolutional financial AI model used for stock market prediction process. The financial AI model is embedded in
the CLEAR-Trade visualization process, which augments a set of fully convolutional
layers, a leaky rectified linear unit layer, global average pooling (GAP)
and a softmax layer at the end of the model for the learning process.}

\label{fig:arch}
\end{figure}

\section{Experiments and Results}
This section explains the experimental setup, the deep learning-driven financial AI model built for performing binary stock market prediction on the S\&P 500 stock market index, and the experimental results for obtaining the efficacy of the CLEAR-Trade visualization in creating interpretable and transparent deep learning-driven financial AI models for stock market prediction.

\begin{figure}[t]
\begin{center}
   \includegraphics[trim = 0cm 1cm 4cm 0cm ,height = 6.5cm,width=1\linewidth]{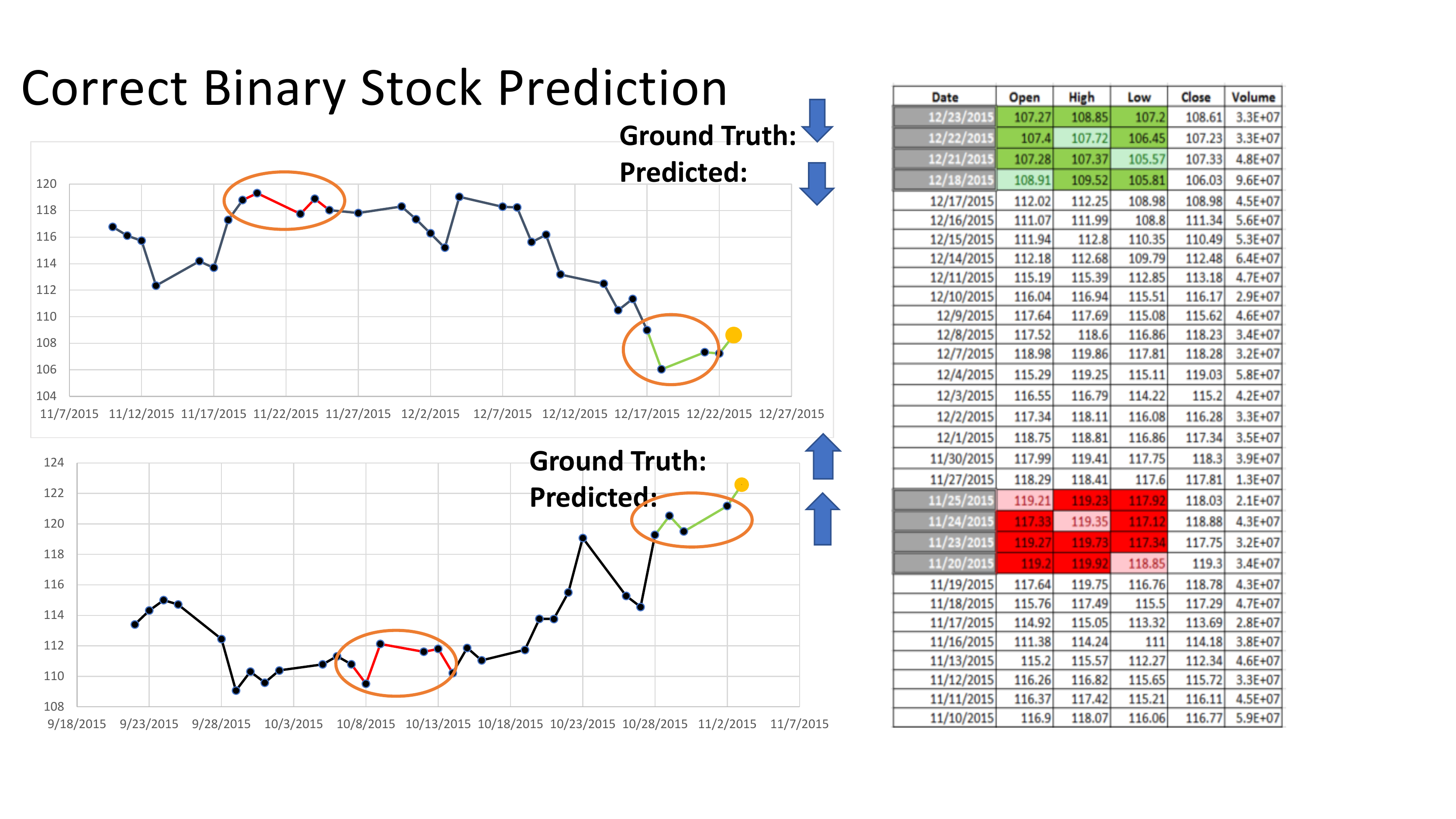}
\end{center}
\vspace{-1cm}

\begin{center}
   \includegraphics[trim = 0cm 2cm 4cm 0cm ,height = 6.5cm,width=1\linewidth]{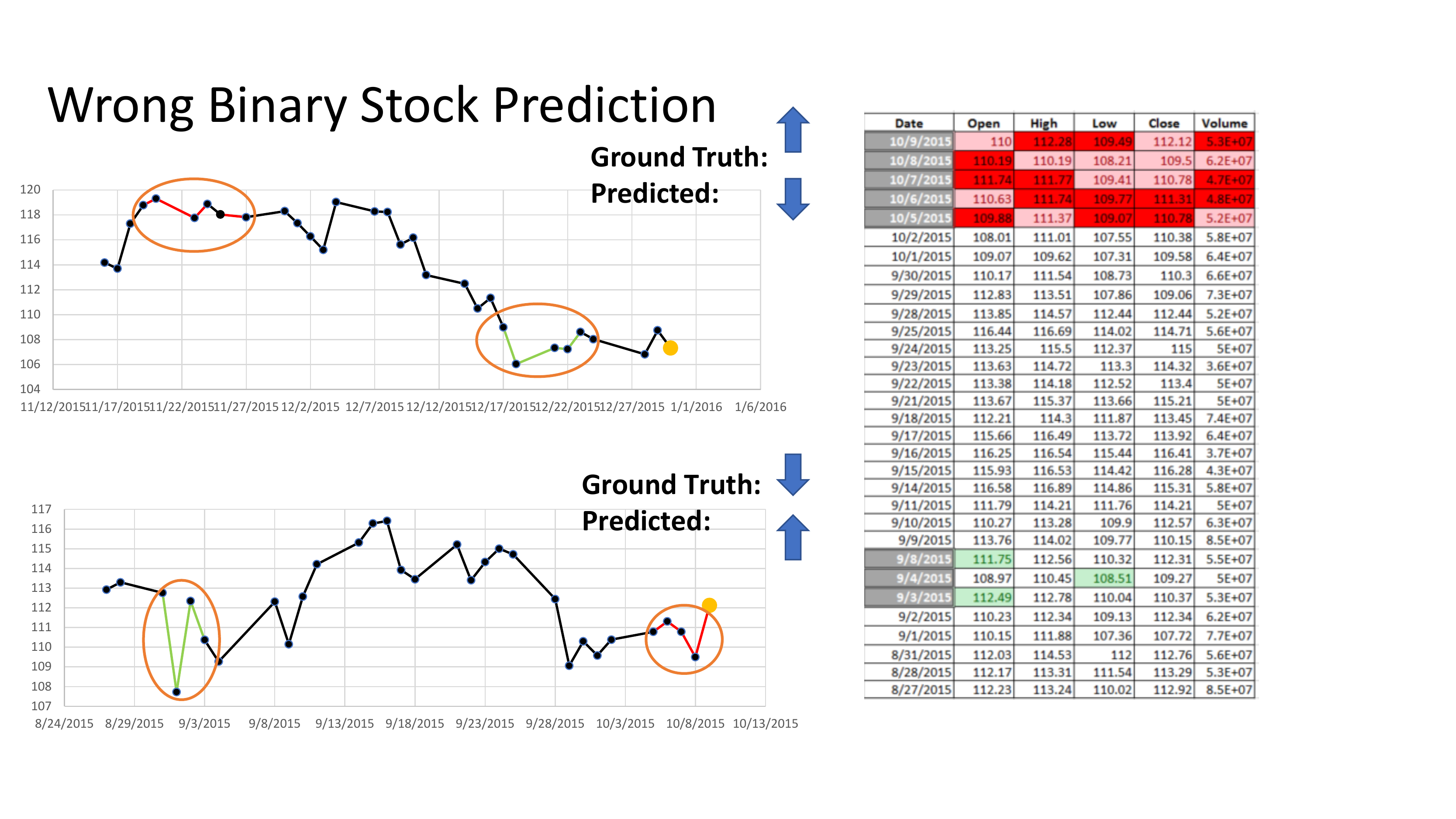}
\end{center}
   \caption{Correctly and mis-classified binary stock market prediction for predicting the S\&P 500 stock market index as visualized and explained using CLEAR-Trade for a deep learning-driven financial AI model trained for experimental purposes. For both cases, the CLEAR-Trade visualizations (left) pinpoint the attentive time windows that are responsible for the particular stock market prediction decision made by the financial AI model (green for index rise and red for index fall). In the CLEAR-Trade visualizations, the thickness of color lines indicate the influence of the attentive time windows on the prediction decision output. On the right hand side, the time-wise stock data information sheet (open, high, low, close, and trade volume) of one of the graphs is shown in the descending order from the day on which decision is being made. Here also, the attentive levels (transparency of color) and the dominant state (green or red) associated with these attentive time windows is shown so that we can better understand why a decision was made.}

\label{fig:results}
\end{figure}

\subsection{Experimental Setup}
For training purposes, we selected the last three years worth of trade data of the S\&P 500 stock market index to train a deep convolutional neural network, shown in Fig.~\ref{fig:arch},  as the deep learning-driven AI financial model used in this study. For preparing this data for training the financial model, we divided the data into 30-day time segments and treated the state (index rise or index fall) on the 31st day as '1' if the index was higher than previous day or '0' if the index was lower. We used 90\% of the data as training set and consider 10\% for evaluation purposes. The trained deep learning-driven financial AI model achieved a prediction accuracy of 61.22\%, though it is important to note that the focus of this paper is on the ability to visualize and understanding the decision-making process of the financial AI model, not in attaining the best possible accuracy, and therefore improving the accuracy of the reference model are plans for future work.

\subsection{Stock Market Prediction Results}
To present the effectiveness of the CLEAR-Trade visualization (as explained in Section~\ref{method}) to enable interpretable deep learning-driven financial AI, we used the trained stock market predictive model and obtained the CLEAR-Trade visualization results as shown in Fig.~\ref{fig:results}. In Fig.~\ref{fig:results}, for both cases (correct and wrong predictions), it can be clearly observed from the CLEAR-Trade visualizations which time windows are most crucial to the decision-making process of the financial AI model for reaching a particular stock market prediction.

Specifically, in the correctly predicted cases, it can be observed that the deep learning-driven financial AI model primarily leveraged the past four days of trade data for correctly predicting whether the S\&P 500 stock market index will rise or fall.  This is intuitive as the past few days are more likely to have a major effect on the index's behavior compared to data from a couple of weeks back. In the case whether the financial AI model gets the stock market prediction incorrect, we can observe that the model primarily leverages trade data for nearly 3 weeks ago to making its decision.  Another observation that can be made in both cases is that in the cases where the stock market prediction is correct, the deep learning-driven financial AI model leverages only open, high and low values to make a decision. This trend if leveraging only open, high, and low values is observed across the majority of correct predictions made by financial AI model. This is again intuitive as unless there is a significant change in trade volume, knowledge of trade volume generally does not have a significant impact on either index rise or fall.  Conversely, it can be seen that when making incorrect decisions, the financial AI model strongly takes into account the trade volume as well, which is not a strong predictive feature as indicate above and as such can incorrectly influence its decisions.  Finally, it can be observed that the confidence of the stock market prediction in this case is low when it makes incorrect predictions, as indicated by the color transparency in the data-sheet, while in correct predictions the confidence of the stock market prediction is high.

Hence, based on the above mentioned observations, it is evident that CLEAR-Trade visualization not only provides a justification for particular stock market prediction decision output, it can also provide considerable insights that financial analysts can taken into account while making trading decisions.

\section{Conclusion}
In this paper, we proposed CLEAR-Trade, a visualization framework that provides insight into the minds of deep learning-driven financial AI models used for stock market prediction by visualizing and explaining the decision-making process of the model. Experiments pertaining to stock market prediction for the S\&P 500 index showed that CLEAR-Trade visualization leads to a higher degree of human interpretability and transparency in predictions made using deep learning-driven financial AI models, hence paving a way for their use in regulatory settings. The proposed visualization approach has tremendous potential to create industry-wide effect by facilitating the use of state-of-the art deep learning models for areas in finance that are under significant regulations.

\section*{Acknowledgments}
This research has been supported by Canada Research
Chairs programs, Natural Sciences Engineering Research
Council of Canada (NSERC), and Canada Foundation for
Innovation (CFI).

{\small
\bibliographystyle{plain}
\bibliography{egbib}
}

\end{document}